\documentclass[letterpaper, 10pt, conference]{ieeeconf} 

\IEEEoverridecommandlockouts                              
\usepackage[T1]{fontenc}
\usepackage[utf8]{inputenc}                              
\usepackage{xcolor}                                      
\usepackage{marvosym}                                    
\usepackage{amsmath,amssymb}                               
\usepackage{graphicx}                                    
\usepackage{algorithm}
\usepackage{algpseudocode}
\usepackage{booktabs}                                    
\usepackage{multirow}    
\usepackage{cuted}
\usepackage[font=small,labelfont=bf]{subcaption}
\usepackage{hyperref}     

\hypersetup{
    colorlinks=true,
    linkcolor=black,
    citecolor=black,
    urlcolor=black,
    pdfborder={0 0 0}
}

\title{\LARGE \bf SA-HGNN: Sample-Adaptive Hyperbolic Graph Neural Network for EEG-Based Depression Recognition}

\author{Yang Li$^{1*}$, Pan Hu$^{1*}$, Yan Zhang$^{1}$, Wenfan Yang$^{1}$, 
Lianbo Guo$^{1}$, and Tao Wu$^{1\dagger}$ 
\thanks{$^{1}$School of Software Engineering, Huazhong University of Science and Technology, Wuhan, Hubei, China}
\thanks{$^{*}$ Equal contribution, $^{\dagger}$ Corresponding author}
}

\begin{document}

\maketitle
\thispagestyle{empty}
\pagestyle{empty}

\begin{abstract}
Graph Neural Networks (GNNs) have been widely used to capture spatial functional connectivity patterns to improve electroencephalography (EEG)-based depression recognition performance. However, the functional connectivity of brain networks in patients with depression exhibits an inherent hierarchical structure, making it difficult to capture accurate connection patterns. To address these issues, this paper proposes a novel model named Sample-Adaptive Hyperbolic Graph Neural Network (SA-HGNN), which aims to accurately extract the authentic hierarchical structure of depression-affected brain networks. Specifically, the proposed model comprises three core modules. First, a Sample-Adaptive Graph Construction module dynamically constructs personalized brain network topologies to capture more complex spatial relationships within the brain network. Second, hyperbolic graph convolution is employed to overcome the representation bottlenecks of Euclidean space, leveraging hyperbolic geometry to precisely capture latent hierarchical relationships within the brain network. Finally, an Attention Pooling module adaptively filters out highly redundant noise channels in EEG signals, effectively mitigating the interference of inherent noise on the authentic hierarchical topology. Extensive experiments on public EEG datasets demonstrate the superior performance of our method across resting-state and task-related paradigms, validating its robustness to noise and efficacy in capturing abnormal functional connectivity patterns in brain networks of patients with depression.

\end{abstract}

\section{INTRODUCTION}
\label{sec:introduction}

Major Depressive Disorder (MDD) has become a leading global health crisis. According to the World Health Organization, over 332 million people suffer from MDD\cite{who2023depressive}, leading to trillions of dollars in annual economic losses \cite{world2025over}. Given the insidious nature of its early symptoms, developing timely and objective automated diagnostic methods is critical. Due to its non-invasiveness, low cost, and high temporal resolution, Electroencephalography (EEG) is widely used to capture dynamic electrophysiological activities for MDD diagnosis \cite{fries2015rhythms}.

Early deep learning approaches, such as Convolutional Neural Networks (CNNs) and Transformers, have shown strong capabilities in extracting spatiotemporal EEG features \cite{liu2022spatial}. However, extensive neurophysiological studies indicate that MDD is fundamentally a disruption of whole-brain functional connectivity rather than a localized lesion, which is closely related to clinical symptoms \cite{kaiser2015large}. Consequently, Graph Neural Networks (GNNs) have rapidly become the mainstream approach, as they can naturally model the brain network and effectively capture these abnormal functional connectivity patterns \cite{bessadok2022graph}.

Focusing on graph structure learning for MDD, recent GNN methods have achieved promising results by explicitly modeling EEG topological structures. For example, MAST-GCN \cite{mast} utilizes adaptive spatial-temporal convolutions to capture dynamic EEG dependencies, while TFAGL \cite{xu2025tfagl} constructs dynamic local-global graphs to simulate regional brain collaborations.

\begin{figure}[t]
  \centering
  \includegraphics[width=0.5\textwidth]{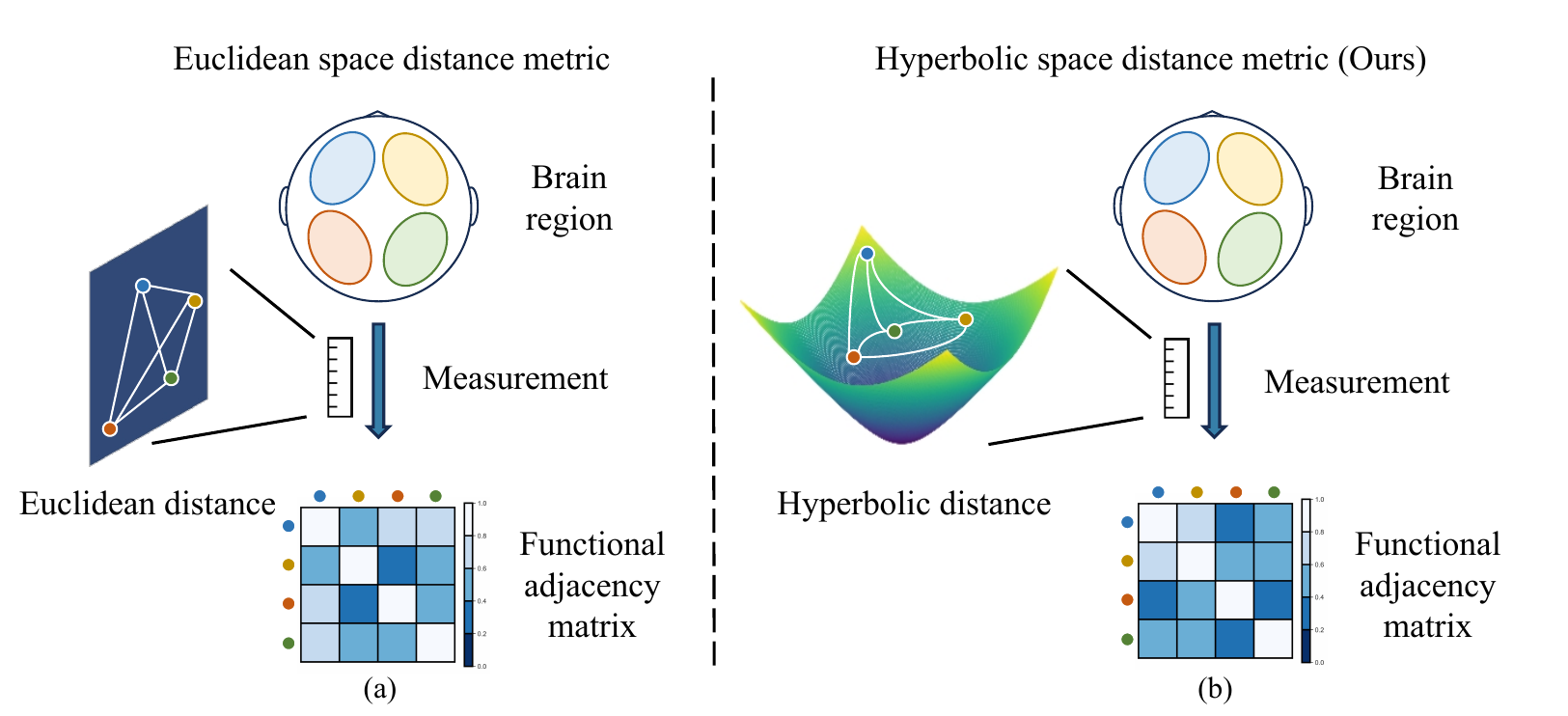} 
  \caption{Measurement methods of brain functional connectivity: (a) Euclidean space metric. (b) Hyperbolic space metric (Ours).}
  \label{fig:gnn_comparison}
  \vspace{-15pt}
\end{figure}

However, human brain functional connectivity is intrinsically hierarchical, featuring a "rich-club" topology where a core set of highly interconnected hub nodes links to peripheral nodes \cite{van2011rich}. Notably, the topology of these hub nodes is severely disrupted in MDD patients \cite{wang2016disorganized}. As illustrated in Fig. \ref{fig:gnn_comparison}, existing GNN methods typically perform computations in Euclidean space. However, this space struggles to accommodate exponentially growing hierarchical tree-like structures, leading to structural distortion and affecting the accurate modeling of the inherent structural hierarchy in brain networks \cite{nickel2017_poincare_embeddings}.

To tackle these challenges, we propose the Sample-Adaptive Hyperbolic Graph Neural Network (SA-HGNN) to model the hierarchical organization of complex brain connectivity. By introducing a hyperbolic space metric, this framework enables hyperbolic graph neural networks to accurately capture latent hierarchical patterns from sample-adaptive topologies. To further extract discriminative features from crucial hub nodes, an attention pooling module is incorporated to filter out redundant noise and preserve the authentic topology of the core subgraph.

\textbf{To the best of our knowledge, this is the first work to adopt Hyperbolic Graph Neural Networks to model the hierarchical functional connectivity of MDD brain networks.} The main contributions of this paper can be summarized as follows:
\begin{itemize}
\item We proposed the Sample-Adaptive Graph Construction (SAGC) module to dynamically construct individualized brain network topologies, overcoming the rigidity of static graphs to provide structural priors.

\item We proposed an architecture integrating Hyperbolic Graph Convolution (HGC) and Attention Pooling (AP) modules to capture hierarchical structures in hyperbolic space while filtering out noisy channels to preserve genuine topologies.

\item We demonstrated through experiments on public EEG datasets that SA-HGNN achieves results under both resting-state and task-related paradigms, outperforming traditional GNNs based on Euclidean space metrics.
\end{itemize}

\section{RELATED WORK}
\label{sec:RELATED_WORK}

\subsection{EEG Based Depression Recognition}

Methods for EEG-based depression recognition typically involve the two key components of feature extraction and classifier design. In terms of feature extraction, existing methods are primarily categorized into handcrafted features and deep learning features. Handcrafted features, such as Power Spectral Density \cite{yang2023depression} are widely used, whereas deep learning methods utilize CNNs to automatically learn discriminative representations from raw EEG signals. Regarding classifier design, research typically models EEG signals from both temporal and spatial perspectives. From a temporal standpoint, models such as CNNs and Transformers are employed to capture the temporal dynamics of EEG signals\cite{yu2025automatic}. From a spatial perspective, graph neural networks regard EEG electrodes as graph nodes to construct brain functional networks, which enables them to effectively model functional connectivity between brain regions\cite{song2018eeg}.

\subsection{Hyperbolic Graph Neural Networks}

Hyperbolic geometry is characterized by a constant negative curvature, where the spatial volume expands exponentially with distance. This spatial property allows it to accommodate nodes whose number grows exponentially with hierarchical depth, making it suitable for modeling graph data with hierarchical structures. Building on this, the classic projection method \cite{nickel2017_poincare_embeddings} employs Poincaré embeddings to model scale-free graphs. Furthermore, Hyperbolic Graph Neural Networks extend traditional message-passing mechanisms to hyperbolic manifolds \cite{chami2019_hgcn_hyperbolicgcn}. Specifically, they map Euclidean node features into hyperbolic space and perform convolutions within the tangent space. By matching the non-Euclidean structure of graph data, such networks can better model hierarchical structures compared with traditional Euclidean models.


Compared with existing methods, the proposed SA-HGNN embeds depressive brain networks into hyperbolic space, enabling a more efficient capture of brain network connectivity patterns with hierarchical features than models that perform computations in the standard Euclidean space.

\section{METHODOLOGY}
\label{sec:methodology}

\begin{figure*}[t]
\centering
\includegraphics[width=0.85\textwidth]{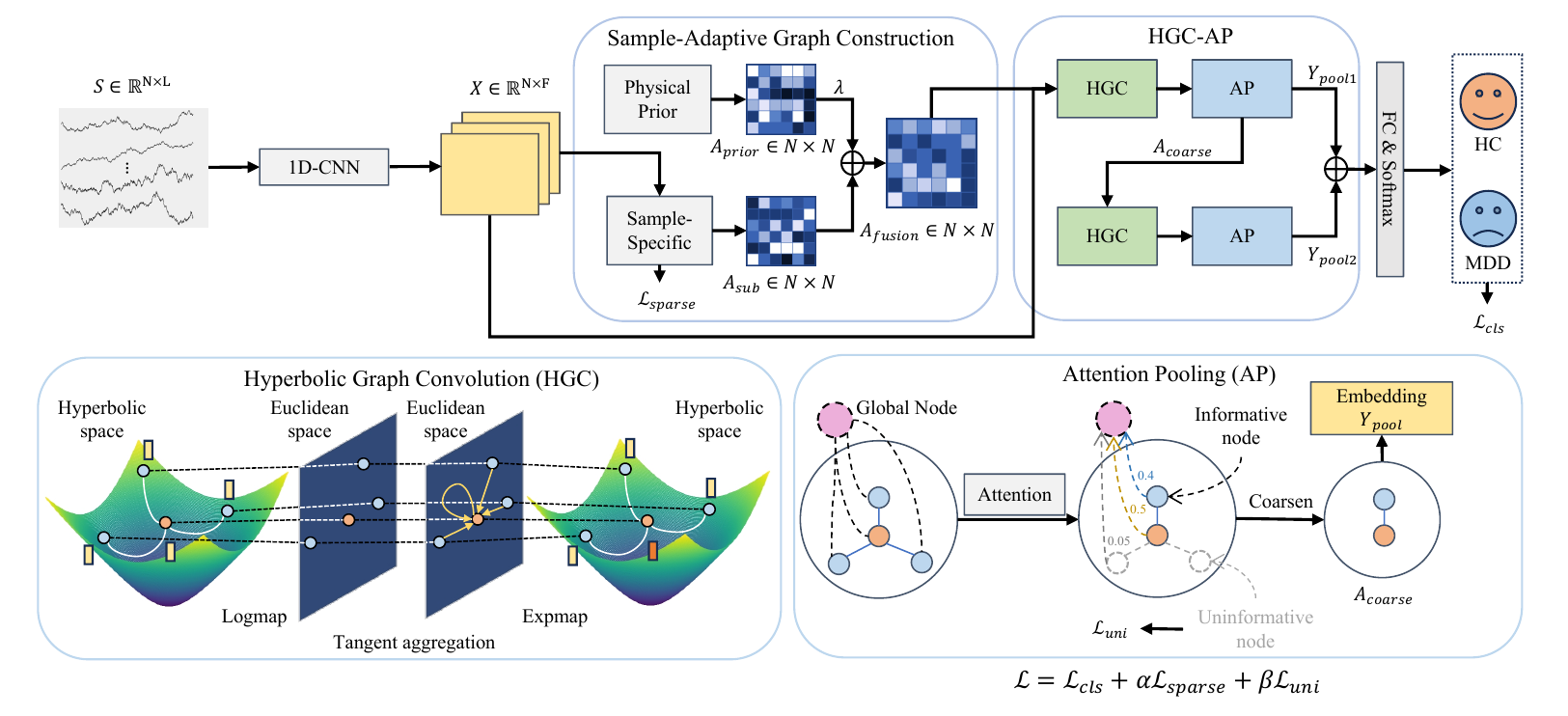}
\caption{Overall architecture of the proposed SA-HGNN model.}
\label{fig:overview}
\end{figure*}

This section presents the SA-HGNN model for EEG-based depression recognition. As illustrated in Fig.~\ref{fig:overview}, the model comprises three core modules: SAGC, HGC, and AP. Specifically, it extracts electrode node features from raw EEG signals, builds sample-adaptive graph topologies, and learns discriminative embeddings via stacked HGC-AP layers for final depression classification.

\subsection{EEG Temporal Feature Extraction}
\label{subsec:eegencoder}

Initially, each EEG sample is defined as a matrix $S \in \mathbb{R}^{N \times L}$, where $N$ is the number of EEG channels and $L$ is the number of time points. To initiate the process, per-channel temporal features are extracted to generate the node feature matrix as \cite{wang2025_hybridgnn_depression_eeg}:
\begin{equation}
X = f(S),
\end{equation}
where $X \in \mathbb{R}^{N \times F}$ is the extracted feature matrix, $f(\cdot)$ denotes the 1D-CNN-based feature extraction function, and $F$ represents the dimension of temporal features extracted for each channel.

\subsection{Sample-Adaptive Graph Construction}
\label{subsec:saa}
The SAGC module fuses physical priors and data-driven correlations to comprehensively model inter-sample connectivity.

\subsubsection{Physical Prior Branch}
To establish a baseline spatial topology, a global physical prior matrix $A_{\text{prior}} \in \mathbb{R}^{N \times N}$ is first constructed from standard 10--20 electrode coordinates:
\begin{equation}
A_{\text{prior}}(i,j) = \min\!\left(1,\max\!\left(0.1,\frac{\delta}{d_{ij}^{2}}\right)\right),
\label{eq:prior}
\end{equation}
where $d_{ij}$ represents the Euclidean distance between two electrodes, and $\delta$ refers to a scaling factor. The diagonal values of this matrix are explicitly set to zero.
\subsubsection{Sample-Specific Correlation Branch}
Furthermore, to derive a data-driven similarity matrix, the extracted features $X$ are first L2-normalized and projected via a linear layer to obtain the transformed features $\overline{X}$. The sample-specific similarity matrix $A_{\text{sim}} \in \mathbb{R}^{N \times N}$ is then computed via their inner product:
\begin{equation}
A_{\text{sim}} = \overline{X}\,\overline{X}^{\top}.
\label{eq:sim}
\end{equation}
Subsequently, to dynamically obtain the sample-specific adjacency matrix $A_{\text{sub}}$, we retain only the positive correlations in $A_{\text{sim}}$ and perform an element-wise multiplication with a learnable mask $M \in \mathbb{R}^{N \times N}$.

\subsubsection{Fusion and Normalization}
To integrate these dual representations, the two matrices are fused using a learnable scalar:
\begin{equation}
A_{\text{fusion}} = A_{\text{sub}} + \lambda A_{\text{prior}},
\label{eq:fusion}
\end{equation}
where $\lambda$ denotes the learnable scalar and $A_{\text{fusion}} \in \mathbb{R}^{N \times N}$ denotes the fused matrix.

Finally, self-loops are added to the fused matrix $A_{\text{fusion}}$, followed by symmetric normalization to obtain the final adjacency matrix $\hat{A}$. To prevent overfitting, L2 regularization is additionally imposed on the SAGC parameters.

\subsection{Hyperbolic Graph Convolution}
\label{subsec:hgc}
Graph convolution operations are conducted within the Poincaré ball model to effectively capture the hierarchical nature of brain networks.

\subsubsection{Mapping Between Euclidean and Hyperbolic Space}
Initially, exponential and logarithmic maps are defined at the origin:
\begin{align}
\exp_0^c(v) &= \tanh(\sqrt{c}\|v\|)\frac{v}{\sqrt{c}\|v\|}, \\
\log_0^c(y) &= \operatorname{arctanh}(\sqrt{c}\|y\|)\frac{y}{\sqrt{c}\|y\|},
\label{eq:log}
\end{align}
where $\exp_0^c$ maps Euclidean features to the hyperbolic space, $\log_0^c$ maps hyperbolic features back to the Euclidean space, $c$ denotes a trainable curvature parameter, $v$ denotes Euclidean vectors, $y$ denotes hyperbolic vectors, and $\|\cdot\|$ denotes the norm operation.
\subsubsection{Hyperbolic Graph Convolution Layer}
Specifically, the input Euclidean features are first projected onto the tangent space via a mapping function and subsequently transformed into the hyperbolic space:
\begin{equation}
X_{\text{hyp}} = \exp_0^c(\phi(X_{\text{euclid}})),
\end{equation}
where $X_{\text{euclid}}$ denotes the input Euclidean features, $X_{\text{hyp}} \in \mathbb{R}^{N \times F}$ denotes the resulting hyperbolic node features, and $\phi(\cdot)$ defines the projection onto the tangent space.

Following this, a linear transformation is applied to the mapped features:
\begin{equation}
H_{\text{lin}} = \exp_0^c\left( W\log_0^c(X_{\text{hyp}}) + b \right),
\label{eq:hyplinear}
\end{equation}
where $W \in \mathbb{R}^{F \times D}$ denotes the learnable weight matrix, $b$ denotes the bias vector, and $H_{\text{lin}} \in \mathbb{R}^{N \times D}$ denotes the transformed hyperbolic features with the hidden dimension $D$.

Neighboring structural information is subsequently aggregated within the tangent space:
\begin{equation}
H_{\text{agg}} = \exp_0^c\!\left( \sum_{j\in\mathcal{N}(i)} \hat{A}_{ij} \log_0^c(H_{\text{lin},j}) \right),
\label{eq:hypagg}
\end{equation}
where $\mathcal{N}(i)$ denotes the neighbor set of node $i$, $\hat{A}_{ij}$ denotes the adjacency weight, and $H_{\text{agg}} \in \mathbb{R}^{N \times D}$ denotes the aggregated features.

Ultimately, the derived hyperbolic features are mapped back to the Euclidean space:
\begin{equation}
Y_{\text{euc}} = \log_0^c(Y_{\text{hyp}}),
\label{eq:hyp_to_eucl}
\end{equation}
where $Y_{\text{hyp}}$ denotes the activated hyperbolic features corresponding to $H_{\text{agg}}$, and $Y_{\text{euc}} \in \mathbb{R}^{N \times D}$ denotes the final output features of the HGC layer.

\subsection{Attention Pooling}
\label{subsec:grepool}
The attention pooling module is designed to systematically filter out irrelevant nodes and select discriminative nodes to generate a global graph embedding.

\subsubsection{Node Scoring via Multi-Head Attention}
First, a learnable global node is introduced to interact with all $N$ EEG channel nodes. By applying multi-head self-attention on the concatenated feature matrix $\tilde{Y} \in \mathbb{R}^{(N+1) \times D}$, we obtain the attention weight tensor $\mathcal{W}_{\text{attn}} \in \mathbb{R}^{H \times (N+1) \times (N+1)}$ across $H$ attention heads. 

For each original EEG node $i$, its importance score $s_i$ is computed by averaging the attention weights from the global node to node $i$ across all heads:
\begin{equation}
s_i = \frac{1}{H} \sum_{h=1}^H W_{i}^{(h)},
\label{eq:node_score}
\end{equation}
where $W_{i}^{(h)}$ denotes the attention weight assigned by the global node to node $i$ in the $h$-th attention head, extracted from $\mathcal{W}_{\text{attn}}$. These individual scores are then concatenated to form the final score vector $S \in \mathbb{R}^N$, which measures the global importance of each EEG node.

\subsubsection{Coarsened Graph and Global Embedding}
To retain the most informative nodes at the $l$-th layer, we define an index set $\mathcal{I}$ comprising the indices of the $K$ nodes with the highest importance scores in $S$. Based on this selected subset, the pooled features $Y_{\text{pool}}^{l+1} \in \mathbb{R}^{K \times D}$ and the coarsened adjacency matrix $A^{l+1} \in \mathbb{R}^{K \times K}$ for the $(l+1)$-th layer are computed as:
\begin{equation}
Y_{\text{pool}}^{l+1} = (Y_{\text{euc}}^l \odot S)_{\mathcal{I}}, \quad A^{l+1} = \hat{A}^l_{\mathcal{I}, \mathcal{I}},
\label{eq:pool_features}
\end{equation}
where $Y_{\text{euc}}^l$ and $\hat{A}^l$ denote the mapped Euclidean node features and the adaptive adjacency matrix at the $l$-th layer, respectively. Here, $\odot$ is element-wise multiplication, $(\cdot)_{\mathcal{I}}$ represents row-wise feature extraction, and $\hat{A}^l_{\mathcal{I}, \mathcal{I}}$ explicitly denotes the principal submatrix extracted by preserving the rows and columns corresponding to $\mathcal{I}$ from the $l$-th layer's graph.

\subsubsection{AP Uniformity Loss}
To mitigate over-smoothing and ensure representational diversity, the features of the discarded nodes $Z_l \in \mathbb{R}^{(N-K) \times D}$ at each layer $l$ are constrained by the Attention Pooling (AP) loss:
\begin{equation}
\mathcal{L}_{\text{uni}} = \frac{1}{L}\sum_{l=1}^{L} \text{KL}\!\left( \text{Softmax}(Z_l W_p) \;\|\; \mathcal{U}_C \right),
\label{eq:uniform_loss}
\end{equation}
where $L$ is the number of layers, $W_p \in \mathbb{R}^{D \times C}$ is a learnable projection matrix, $\text{KL}$ denotes the Kullback-Leibler divergence, and $\mathcal{U}_C$ signifies the target uniform distribution over $C$ classes.

\subsection{Classifier and Loss Function}
\label{subsec:classifier}
To yield the final prediction, the global embeddings $g_l$ extracted from all $L$ architectural layers are cumulatively summed to establish the fusion graph embedding $g_{\text{fusion}} \in \mathbb{R}^{D}$. A subsequent linear layer generates the prediction logits directly from this fusion embedding.

For model optimization, standard cross-entropy loss is adopted as the primary classification objective:
\begin{equation}
\mathcal{L}_{\text{cls}} = -\frac{1}{B}\sum_{i=1}^{B} \log \frac{\exp(\hat{y}_{i,c_i})}{\sum_{j=1}^{C} \exp(\hat{y}_{i,j})},
\label{eq:loss_cls}
\end{equation}
where $\mathcal{L}_{\text{cls}}$ denotes the classification loss, $B$ denotes the batch size, $\hat{y}$ denotes the prediction logits, and $c_i$ denotes the true ground-truth label.

In addition, a sparse loss is computed specifically for the SAGC module parameters to encourage simpler graph topologies:
\begin{equation}
\mathcal{L}_{\text{sparse}} = \|M\|_F + \|\lambda\|_2,
\label{eq:loss_sparse}
\end{equation}
where $\mathcal{L}_{\text{sparse}}$ denotes the sparse loss, $\|\cdot\|_F$ denotes the standard Frobenius norm, and $\|\cdot\|_2$ denotes the L2 norm.

Finally, the three independent loss terms are jointly combined using balancing hyperparameters:
\begin{equation}
\mathcal{L} = \mathcal{L}_{\text{cls}} + \alpha \mathcal{L}_{\text{sparse}} + \beta \mathcal{L}_{\text{uni}},
\label{eq:total_loss}
\end{equation}
where $\mathcal{L}$ denotes the total loss, and $\alpha$ and $\beta$ represent hyperparameters strictly tuned for effective loss balancing.

\section{EXPERIMENTS}
\subsection{Datasets and Preprocessing}
\label{subsec:dataset}

We evaluate the proposed model on the publicly available HUSM dataset~\cite{mumtaz2017wavelet}, which contains EEG recordings from 34 patients with MDD and 30 healthy controls (HCs). All signals were recorded at 256~Hz using 19 electrodes placed according to the international 10--20 system. 

\subsubsection{EEG Data Types}
The dataset contains two types of EEG data: resting-state EEG (HUSM-Rest) and task-related EEG (HUSM-Task). HUSM-Rest reflects spontaneous brain activity without external cognitive tasks, while HUSM-Task captures stimulus-evoked responses during task execution. 

\subsubsection{Preprocessing Pipeline}
Following standard EEG preprocessing and artifact rejection procedures \cite{seal2021deprnet}, the final HUSM-Rest dataset consists of 3726 MDD and 3588 HC segments, while the HUSM-Task dataset comprises 13200 MDD and 11200 HC samples.
\begin{table}[H]
  \centering
  \caption{Configuration of Hyperparameters}
  \label{tab:hyperparameters}
  \resizebox{\linewidth}{!}{
    \begin{tabular}{l l c c}
      \toprule
      Hyperparameter & Description       & HUSM-Rest & HUSM-Task \\
      \midrule
      $bs$           & Batch size        & 32        & 32        \\
      $lr$           & Learning rate     & 0.001     & 0.001     \\
      $r$            & AP pooling ratio  & 0.6       & 0.3       \\
      $\beta$      & Uniform loss weight & 0.0005  & 0.01      \\
      $\delta$       & Distance scaling factor & 600   & 700       \\
      $\alpha$            & Matrix sparsification weight & $10^{-5}$ & $10^{-5}$ \\
      \bottomrule
    \end{tabular}%
  }
\end{table}
\subsection{Implementation Details}

\begin{table*}[t]
  \vspace{6pt}
  \centering
  \caption{Comparison with Baseline Methods on the HUSM-Rest and HUSM-Task Datasets}
  \label{tab:baseline_combined}
  \resizebox{0.9\textwidth}{!}{%
  \begin{tabular}{lccccccccc}
    \toprule
    \multirow{2}{*}{Model} & \multicolumn{4}{c}{HUSM-Rest} & \multicolumn{4}{c}{HUSM-Task} \\
    \cmidrule(lr){2-5} \cmidrule(lr){6-9}
    & Accuracy (\%) & F1 Score (\%) & Precision (\%) & Recall (\%) & Accuracy (\%) & F1 Score (\%) & Precision (\%) & Recall (\%) \\
    \midrule
    SDGCN~\cite{cui2023multiview}        & 87.00 & 85.86 & 88.15 & 84.63 & 86.38 & 87.70 & 83.70 & 94.62 \\
    GCBNet~\cite{zhang2019gcb}          & 90.91 & 90.64 & 91.49 & 90.10 & 93.87 & 94.27 & \textbf{92.35} & 97.21 \\
    DGCNN~\cite{song2018eeg}           & 92.23 & 91.14 & 93.77 & 89.40 & 93.39 & 93.30 & 91.47 & 96.43 \\
    LGGNet~\cite{ding2023lggnet}         & 93.09 & 92.66 & 92.99 & 93.01 & 92.93 & 93.49 & 90.78 & 97.23 \\
    RGNN~\cite{zhong2020eeg}           & 93.29 & 93.10 & \textbf{95.79} & 91.03 & 90.05 & 90.32 & 85.63 & 96.32 \\
    GraphSleepNet~\cite{jia2020graphsleepnet} & 93.42 & 93.48 & 93.89 & 93.46 & 91.64 & 92.18 & 90.37 & 95.78 \\
    DCGNN~\cite{xiao2025dynamical}      & 91.61 & 90.61 & 93.91 & 88.08 & 92.32 & 93.06 & 89.46 & 98.43 \\
    \textbf{SA-HGNN (Ours)}             & \textbf{95.24} & \textbf{95.77} & 95.44 & \textbf{96.19} & \textbf{94.26} & \textbf{94.69} & 91.73 & \textbf{98.70} \\
    \bottomrule
  \end{tabular}%

  }
\end{table*}
To ensure reliable and reproducible results, we performed subject-wise 10-fold cross-validation following standard protocols. To prevent information leakage, all data from each subject were exclusively assigned to either the training set or the test set, avoiding overfitting caused by subject-specific information.

Four commonly used evaluation metrics in relevant research were employed to comprehensively assess the depression recognition performance: Accuracy (ACC), Recall (REC), Precision (PRE), and F1 score (F1). 

Experiments were conducted on an NVIDIA GeForce RTX 4090 GPU, and a series of critical hyperparameters were tuned to optimize the model performance. The final optimal configurations are summarized in Table~\ref{tab:hyperparameters}.

\subsection{Compared Methods}

To comprehensively evaluate the proposed SA-HGNN, we compare it with seven classic graph-based models: DCGNN~\cite {xiao2025dynamical}, GraphSleepNet~\cite {jia2020graphsleepnet}, LGGNet~\cite {ding2023lggnet}, DGCNN~\cite {song2018eeg}, SDGCN~\cite {cui2023multiview}, RGNN~\cite {zhong2020eeg}, and GCBNet~\cite {zhang2019gcb}. All adopt graph convolutions to capture spatial relationships between EEG channels.
\section{RESULTS AND DISCUSSION}
\subsection{Comparison of Depression Recognition Performance}
Table \ref{tab:baseline_combined} presents the performance comparison of all methods on the HUSM-Rest and HUSM-Task datasets. The proposed SA-HGNN model achieves state-of-the-art performance across most metrics.

Specifically, on the HUSM-Rest dataset, SA-HGNN yields an accuracy of 95.24\% and an F1 score of 95.77\%, outperforming the second-best baseline GraphSleepNet by margins of 1.82\% and 2.29\%, respectively. Notably, our method also achieves the highest recall of 96.19\%, exceeding GraphSleepNet by 2.73\%.

On the HUSM-Task dataset, SA-HGNN continues to demonstrate competitive performance with an accuracy of 94.26\% and an F1 score of 94.69\%, exceeding the second-best baseline GCBNet by 0.39\% and 0.42\%, respectively. Additionally, it achieves the highest recall of 98.70\%, outperforming the second-best method DCGNN by 0.27\%.

Overall, SA-HGNN achieves consistent high recall and F1 scores across both datasets, effectively minimizing false negatives, which is critical for clinical MDD screening. These results validate our proposed architecture: the SAGC and HGC modules capture latent hierarchical brain networks, while the AP mechanism filters out noise to support stable performance.

\subsection{Ablation Study}
To validate the individual and synergistic effects of the three core modules in our SA-HGNN model, we perform a comprehensive ablation study evaluating seven variants. Specifically, we denote the HGC, SAGC, and AP modules as A, B, and C, respectively.

The results of the ablation study are presented in Table \ref{tab:ablation_combined}.
The ablation results show that single-component variants A, B, C achieve stable baseline performance, while pairwise combinations AB, AC, BC bring consistent gains, showing strong synergies between modules. 
The full model ABC achieves the best overall performance, with 95.24\% accuracy and 95.77\% F1 in the HUSM-Rest, and 94.26\% accuracy and 94.69\% F1 in the HUSM-Task, outperforming all other variants. 
This verifies that A, B, and C are mutually reinforcing and together form an effective framework for brain signal analysis.

\subsection{Parameter Sensitivity Analysis}
\begin{figure}[h!]
    \centering
    \vspace{4pt}
    \begin{subfigure}[b]{0.235\textwidth}
        \centering
        \includegraphics[width=\linewidth]{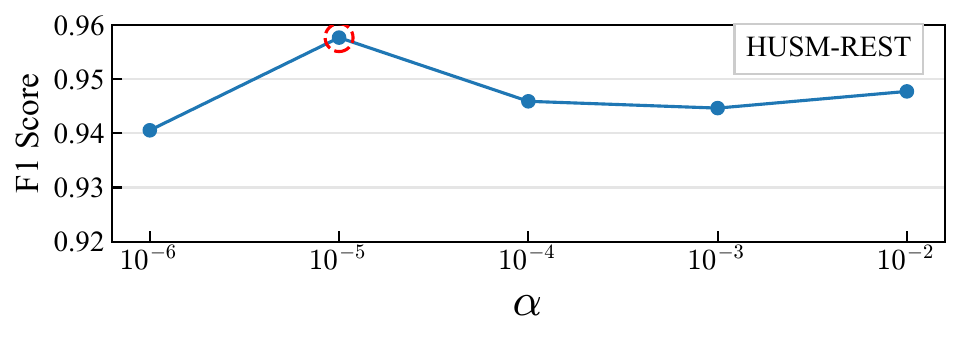}
        \label{fig:sub-1}
    \end{subfigure}
    \begin{subfigure}[b]{0.235\textwidth}
        \centering
        \includegraphics[width=\linewidth]{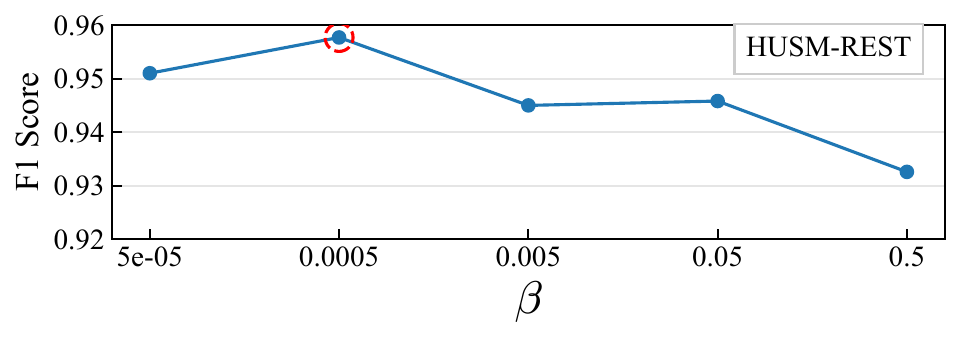}
        \label{fig:sub-2}
    \end{subfigure}
    \vspace{-15pt}

    \begin{subfigure}[b]{0.235\textwidth}
        \centering
        \includegraphics[width=\linewidth]{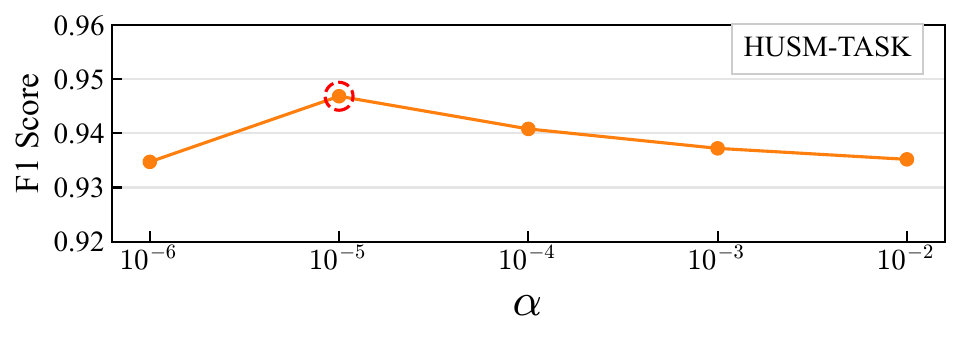}
        \label{fig:sub-3}
    \end{subfigure}
    \begin{subfigure}[b]{0.235\textwidth}
        \centering
        \includegraphics[width=\linewidth]{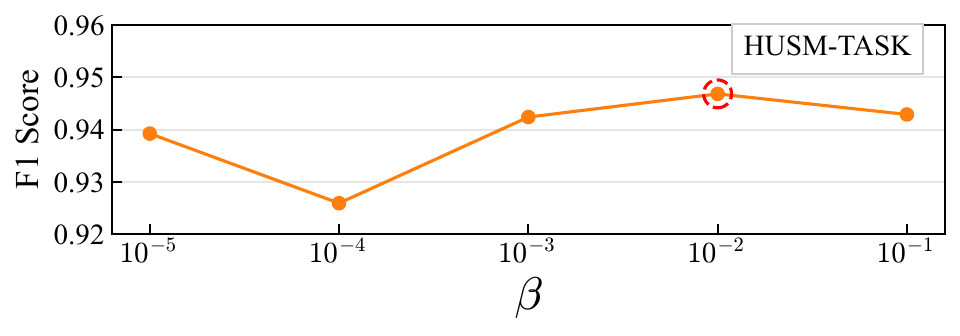}
        \label{fig:sub-4}
    \end{subfigure}
    \vspace{-15pt}
    \caption{Hyperparameter optimization for the matrix sparsification weight $\alpha$ and uniform loss weight $\beta$ on the HUSM-Rest and HUSM-Task datasets.}
    \label{fig:four_images_grid}
    \vspace{-15pt}
\end{figure}

To evaluate model robustness and the impact of loss balancing, hyperparameter sensitivity is investigated on both HUSM-Rest and HUSM-Task datasets. Detailed optimization results for matrix sparsification weight $\alpha$ and uniform loss weight $\beta$ are illustrated in Fig.~\ref{fig:four_images_grid}.

Specifically, the model achieves its optimal F1 score at $\alpha = 10^{-5}$ on both datasets. Furthermore, the performance exhibits minor fluctuations and remains at a high level when $\alpha$ varies from $10^{-6}$ to $10^{-2}$ in steps of powers of 10. As for the uniform loss weight $\beta$, on the HUSM-Rest dataset, the model reaches its peak F1 score at $\beta = 0.0005$, maintaining stable performance when $\beta$ varies from $0.0005$ to $0.05$ in steps of powers of 10. Similarly, on the HUSM-Task dataset, the optimal value is $\beta = 10^{-2}$, with the performance remaining stable and competitive when $\beta$ varies from $10^{-3}$ to $10^{-1}$ in steps of powers of 10. The fact that SA-HGNN maintains stable and competitive results across such discrete hyperparameter evaluation steps clearly demonstrates its strong robustness.
\begin{table*}[t]
  \vspace{6pt}
  \centering
  \caption{Experimental Results of Ablation Study with Different Modules on HUSM-Rest and HUSM-Task Datasets}
  \label{tab:ablation_combined}
  \resizebox{0.9\textwidth}{!}{
  \begin{tabular}{lccccccccc}
    \toprule
    \multirow{2}{*}{Setting} & \multicolumn{4}{c}{HUSM-Rest} & \multicolumn{4}{c}{HUSM-Task} \\
    \cmidrule(lr){2-5} \cmidrule(lr){6-9}
    & Accuracy (\%) & F1 Score (\%) & Precision (\%) & Recall (\%) & Accuracy (\%) & F1 Score (\%) & Precision (\%) & Recall (\%) \\
    \midrule
    A   & 93.07 & 92.82 & 94.45 & 91.79 & 92.44 & 92.71 & 90.14 & 96.61 \\
    B   & 93.08 & 92.38 & 93.74 & 91.64 & 92.14 & 92.04 & 89.80 & 95.16 \\
    C   & 92.79 & 92.80 & 91.72 & 94.46 & 92.01 & 92.21 & 88.98 & 96.15 \\
    AB  & 93.25 & 92.71 & 93.76 & 92.25 & 92.72 & 92.89 & 90.27 & 96.66 \\
    AC  & 93.25 & 93.75 & 94.95 & 93.11 & 92.64 & 92.75 & 90.27 & 95.94 \\
    BC  & 93.40 & 94.30 & 94.23 & 95.28 & 92.69 & 92.65 & 90.00 & 95.93 \\
    ABC & \textbf{95.24} & \textbf{95.77} & \textbf{95.44} & \textbf{96.19} & \textbf{94.26} & \textbf{94.69} & \textbf{91.73} & \textbf{98.70} \\
    \bottomrule
  \end{tabular}
  }
  \vspace{-3pt}
\end{table*}

\section{VISUALIZATION AND DISCUSSION}
\begin{figure}[H]
    \centering
    \begin{subfigure}{0.23\textwidth}
        \centering
        \includegraphics[width=\linewidth]{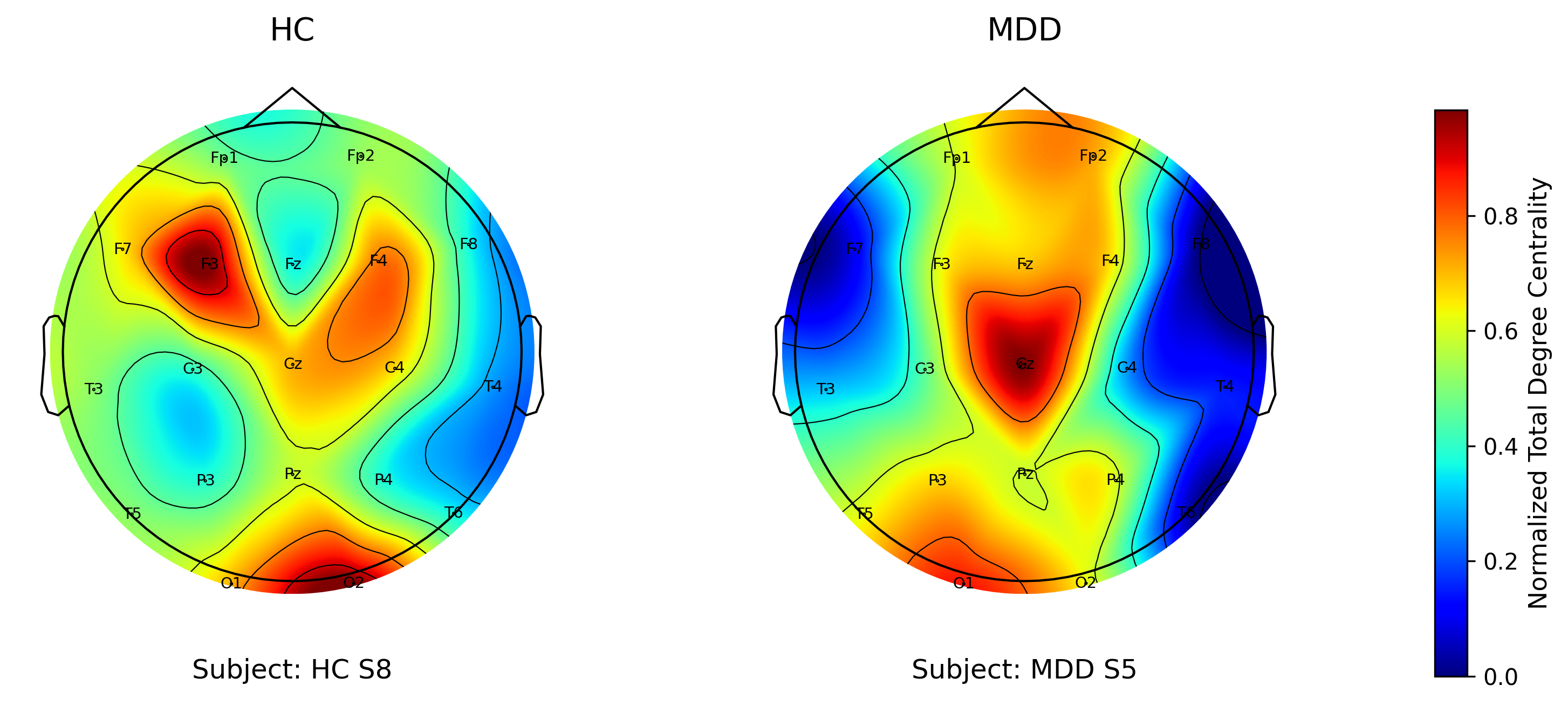}
        \caption{HUSM-REST}
        \label{fig:map_rest}
    \end{subfigure}\hfill
    \begin{subfigure}{0.23\textwidth}
        \centering
        \includegraphics[width=\linewidth]{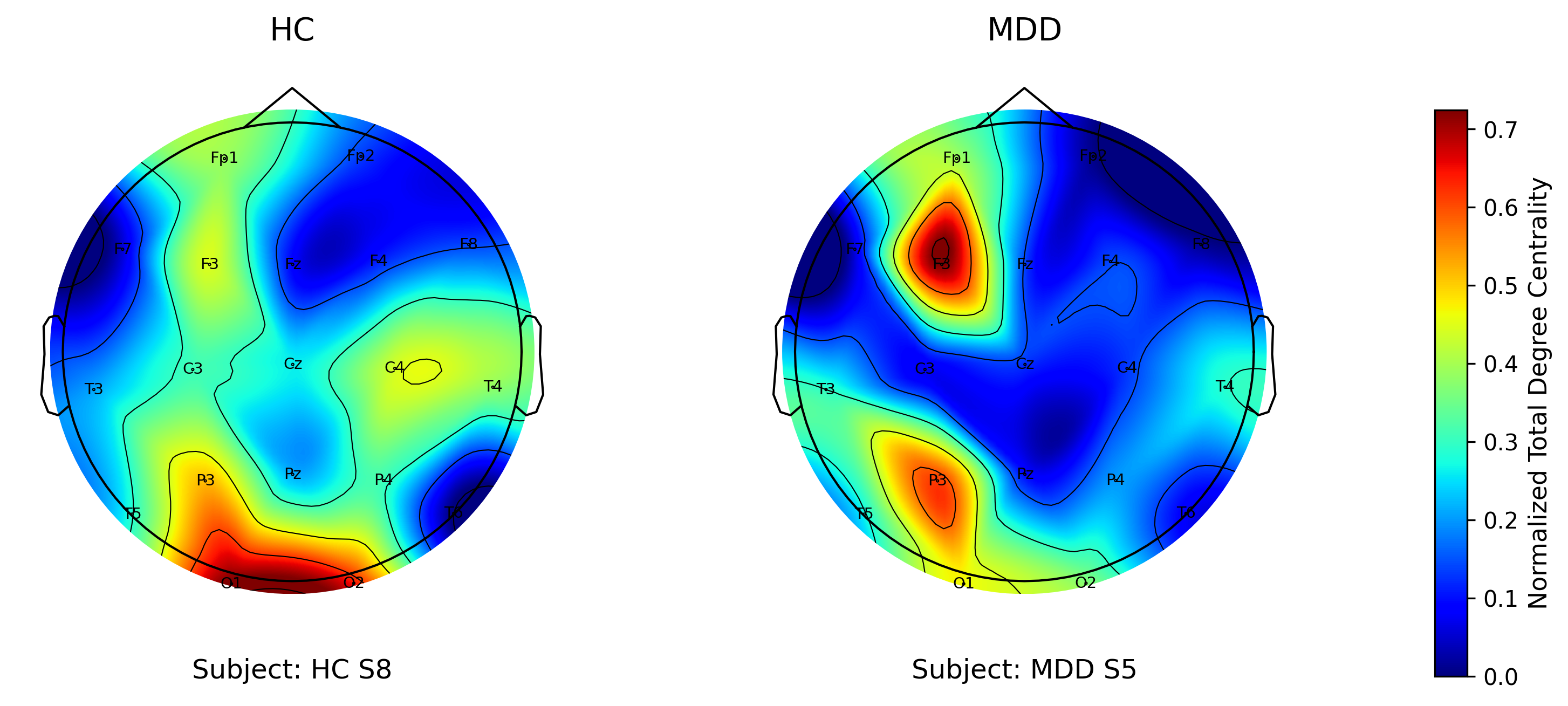}
        \caption{HUSM-TASK}
        \label{fig:map_task}
    \end{subfigure}
    \caption{Topomaps of $A_{\text{fin}}$ for different datasets.}
    \label{fig:A_sub_topomaps}
\end{figure}

To demonstrate the interpretability of the brain connectivity patterns captured by SA-HGNN, we visualized the normalized total degree centrality derived from the learned adjacency matrix. As shown in the Fig.~\ref{fig:A_sub_topomaps}. Compared with healthy controls, MDD subjects show hyperconnectivity in midline central regions during resting states and notable alterations in the left frontal–parietal regions during tasks. These patterns are consistent with clinical findings, where midline hyperconnectivity is associated with default mode network overactivation \cite{kaiser2015large}, and left frontoparietal alterations are linked to emotional–cognitive deficits \cite{disner2011neural}. This result indicates that the proposed method can capture neurobiological biomarkers.

\section{CONCLUSION}
\label{sec:conclusion}

In this article, we propose the Sample-Adaptive Hyperbolic Graph Neural Network (SA-HGNN) for EEG-based depression recognition. To accurately capture the latent hierarchical abnormal patterns in the brain networks of patients with depression, the Sample-Adaptive Graph Construction module is designed to construct personalized topologies by dynamically fusing physical priors and individual feature correlations. Furthermore, the Hyperbolic Graph Convolution and Attention Pooling modules are jointly utilized to embed these hierarchical relationships in hyperbolic space and extract noise-suppressed global graph features. Finally, evaluations of the public HUSM datasets under both resting-state and task-related paradigms demonstrate that SA-HGNN outperforms relevant baseline methods with significant improvements, validating the overall effectiveness of our proposed method.

\addtolength{\textheight}{-12cm}

\bibliography{reference}

\end{document}